\documentclass[runningheads]{llncs}
\usepackage{url}
\usepackage{times}
\usepackage{epsfig}
\usepackage{graphicx}
\usepackage{amsmath}
\usepackage{amssymb}
\usepackage{booktabs}
\usepackage{subfigure}
\usepackage{epstopdf}
\usepackage{comment}
\usepackage{color}
\usepackage{url}
\usepackage{xcolor}
\usepackage{soul}
\usepackage[small]{caption}
\usepackage{makecell}
\usepackage{array}
\usepackage{algorithm}
\usepackage{algorithmic}
\usepackage{bbding}

\graphicspath{{figures/}}
\begin{document}
\title{InSphereNet: a Concise Representation and Classification Method for 3D Object}
%
%
\author{Hui Cao \and
Haikuan Du \and
Siyu Zhang \and
Shen Cai\inst{(}\Envelope\inst{)}}

\authorrunning{Cao et al.}
%
\institute{School of Computer Science and Technology, Donghua University, Shanghai, China\\
\email{\{ch123ui, Hanson\_du, akirazure, hammer\_cai\}@163.com}}
%
\maketitle 
\begin{abstract}
In this paper, we present an InSphereNet method for the problem of 3D object classification. Unlike previous methods that use points, voxels, or multi-view images as inputs of deep neural network (DNN), the proposed method constructs a class of more representative features named infilling spheres from signed distance field (SDF). Because of the admirable spatial representation of infilling spheres, we can not only utilize very fewer number of spheres to accomplish classification task, but also design a lightweight InSphereNet with less layers and parameters than previous methods. Experiments on ModelNet40 show that the proposed method leads to superior performance than PointNet and PointNet++ in accuracy. In particular, if there are only a few dozen sphere inputs or about 100000 DNN parameters, the accuracy of our method remains at a very high level (over 88\%). This further validates the conciseness and effectiveness of the proposed InSphere 3D representation.

\end{abstract}
%

\section{Introduction}
After intensive research in recent years, convolutional neural networks (CNN) is widely used in many areas, such as computer vision, multimedia, and so on. Despite the great success in detection, recognition, segmentation, and classification tasks for 2D images, the use of deep learning on 3D data remains a big challenge because of the sparsity of most 3D data. To the task of 3D object classification, the commonly available datasets are ModelNet \cite{Wu20153DSA} and ShapeNet \cite{Yi2016ASA} in which each object has a complete CAD model. Thus the previous works can transform 3D models into multi-view images, voxels, or point clouds which are then fed into convolutional neural networks. For example, 2D CNN based MVCNN \cite{Su2015MultiviewCN} recognizes 3D shapes from a collection of their rendered views on 2D images which lack of explicit 3D geometric information. VoxNet \cite{Maturana2015VoxNetA3} represents a 3D shape with a volumetric occupancy grid and trains a 3D CNN to perform classification on voxels. However, volumetric CNNs typically have low resolutions (e.g. $32 \times 32 \times 32$) due to computationally expensive 3D convolutions and therefore have difficulty processing fine object models. PointNet \cite{Qi2017PointNetDL} is the first method to apply deep learning directly on points. Though achieving record-breaking results, it is unable to extract local features and its inputs should contain enough points to cover the surface of the object.

In addition, signed distance field (SDF) becomes another popular choice for 3D shape representation. As the SDF value of a spatial point stands for the distance between this point and its nearest object surface, the point whose SDF value is equal to zero lies on the surface. Every point on the exterior of the surface is considered positive distance and any point inside the mesh stores a negative distance. Some fusion methods \cite{Curless1996AVM,Newcombe2011KinectFusionRD} use a truncated SDF (TSDF) to reconstruct a single 3D model from noisy depth maps. Voxel-based SDF representations have been extensively used for 3D shape learning\cite{Zeng20163DMatchLL,Dai2016ShapeCU,Stutz2018Learning3S} and 3D shape completion \cite{Park2019DeepSDFLC}, but their use of discrete voxels is expensive in memory. Although SDF is capable to express the shape of any 3D object, it has not been used for 3D object classification to the best of our knowledge. The main reason why SDF is difficult to be applied to 3D object classification is that it is a dense field with 3D position and SDF value. It suffers from the same problems as voxels.

\begin{figure}[tbh]
\centering
\includegraphics[width=\textwidth]{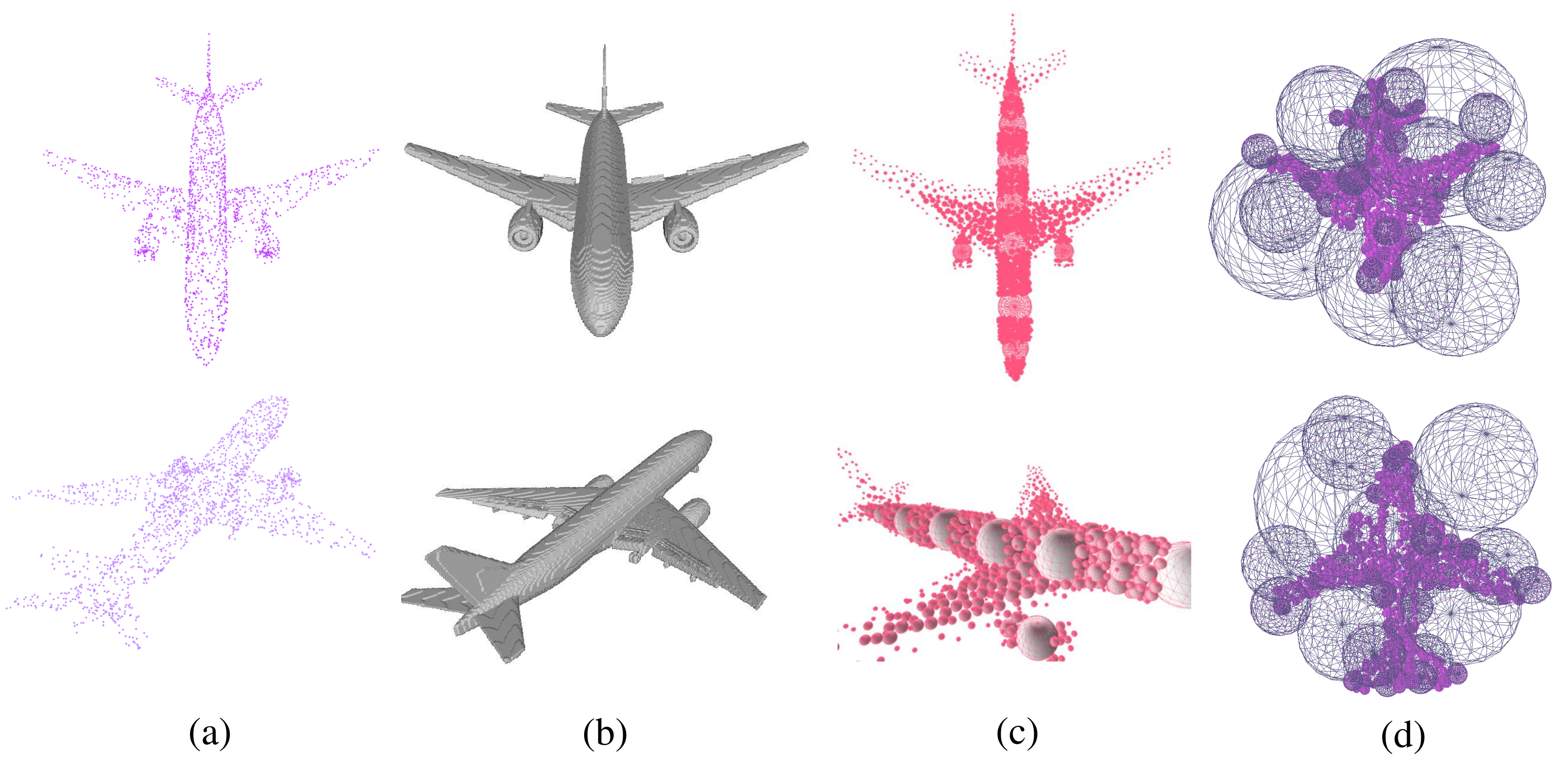}
\caption{Four different primitive representations of an airplane model. (a) 2048 surface points; (b) voxels with the resolution ${512^3}$; (c) 1024 interior infilling spheres; (d) 1024 exterior infilling spheres}
\label{fig:airplane_model_four_forms}
\centering
\end{figure}

In this paper, we propose infilling spheres extracted from the SDF to represent a complete 3D object. For each voxel, we can construct a sphere with its 3D coordinates as the sphere's center and its SDF value as the sphere's radius. A number of spheres (e.g. 64-1024) are selected to represent the object which we name infilling spheres. Intuitively, space infilling spheres are more informative and representative than isolated surface points for 3D objects because a surface point is just equivalent to a sphere with a radius of zero at the specific locations (surface) while a sphere can be anywhere with any size. Fig. \ref{fig:airplane_model_four_forms} shows an airplane model represented by four different primitives which are point clouds, voxels, interior infilling spheres and exterior infilling spheres separately. The proposed infilling spheres representation is more concise and effective than other representation methods, especially with a few of primitives.

Specifically, we first normalize 3D objects into a unit size and voxelize them with a high resolution of $512 \times 512 \times 512$. Then we compute the SDF value of each voxel with which we can construct a sphere by using its position as the sphere's center and its SDF value as the sphere's radius. Subsequently, only a number of infilling spheres are constructed according to three criteria we will introduce later. After that, the infilling spheres with four-dimensional vectors are fed into a lightweight PointNet network architecture.  Experiments on ModelNet40 verify the representation power of the proposed method. If there are only a few dozen sphere inputs or about 100000 DNN parameters, the accuracy of our method remains at a very high level.

The contributions of our work can be summarized as follows:

\begin{itemize}
\item We propose a novel 3D shape representation using infilling spheres which is geometrically intuitive and meaningful.
\item The number of infilling spheres can sharply decreases without obvious decrease of classification accuracy. 
\item The network architecture can be lightweight without obvious decrease of classification accuracy.
\end{itemize}

\section{Related Works}
Traditionally, hand-crafted features for point clouds can be divided into two categories: \emph{intrinsic} and \emph{extrinsic}. Intrinsic descriptors \cite{Rustamov2007LaplaceBeltramiEF,Sun2009ACA,Shah20133DDivAN} treat the 3D shape as a manifold, while extrinsic descriptors \cite{Rusu2008AligningPC,Rusu2009FastPF,Ling2007ShapeCU} are usually extracted from coordinates of the shape in 3D space.

In the deep learning era, approaches to 3D object classification keep evolving at a fast pace. Volumetric CNNs have been adopted by pioneers on this specific task. \cite{Wu20153DSA} proposes to convert depth maps to volumetric representations, and then utilize a convolutional deep belief network to recognize object categories. VoxNet \cite{Maturana2015VoxNetA3} represents point clouds with a volumetric occupancy grid and trains a 3D CNN to accomplish classification on voxels. However, volumetric CNNs typically have low resolutions due to sparsity of points and computationally expensive 3D convolutions and therefore have difficulty processing very large point clouds.

Another family of methods\cite{Su2015MultiviewCN,Qi2016VolumetricAM} classify multi-view 2D images captured from the circular observation of 3D shapes with state-of-the-art 2D CNNs and achieve record-breaking results. However, they are not geometrically intuitive and cannot easily be extended to other 3D tasks such as part segmentation. 

To circumvent these issues, PointNet \cite{Qi2017PointNetDL} directly consumes point clouds with a simple yet efficient network. It is also robust to inputs perturbation and corruptions. Nonetheless, it only considers global features and ignores local neighborhood information, making it not suitable to fine-grained pattern and complex scenes. Instead of working on individual points, PointNet++ \cite{Qi2017PointNetDH} introduces a hierarchical neural network that applies PointNet recursively on several group points in different levels. Consequently, features from multiple scales could be extracted hierarchically. One implicit drawback of this family of methods is that they need a sufficient number of points to cover the whole surface. CNNs are very efficient for processing data representation which have a grid structure. But point clouds usually do not have grids thus makes it hard to learn local information. A simple method to overcome this problem is constructing neural networks on graphs \cite{Bruna2013SpectralNA,wang2019learning,yu2018joint}. DGCNN \cite{Wang2018DynamicGC} exploits local geometric structures by constructing a local neighborhood graph and applying graph convolutions on the connecting edges between points. This approach fails to take directions and other information into account, which is also essential to 3D object recognition.

Different from above 3D representation methods, the methodology we adopted in this paper is to explore another more concise way, which can express geometric models from coarse to fine. After performing a series of construction operations on the SDF of an object, the proposed infilling spheres can work well for classification task, even if the number of spheres is small or a lightweight network is adopted. 

\section{Our Approach}
\begin{figure}[t]
\centering
\includegraphics[width=\textwidth]{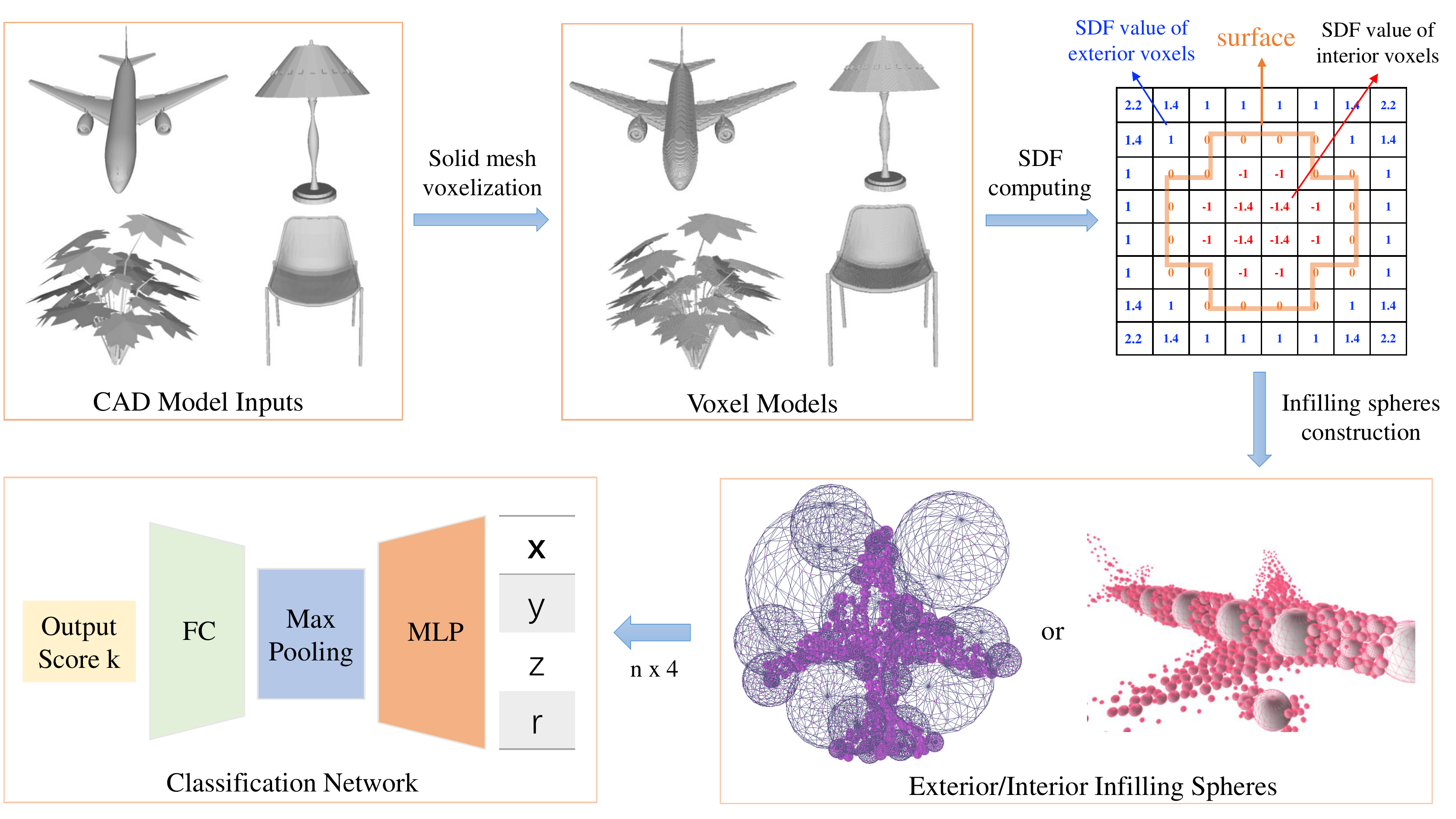}
\caption{The flow chart of InSphereNet}\label{fig:flow}
\centering
\end{figure}

This section describes the proposed approach in detail. Firstly, we voxelize the 3D model with a high resolution of $512 \times 512 \times 512$. Secondly, the SDF value of each voxel within an external sphere is computed. Thirdly, a number of voxels with larger SDF values are selected according to three criteria. Finally, positions and radii of selected infilling spheres are fed into the classification network. The detailed workflow is illustrated in Fig. \ref{fig:flow}.

\subsection{Mesh Voxelization}
\begin{figure}[t]
\centering
\includegraphics[width=\textwidth]{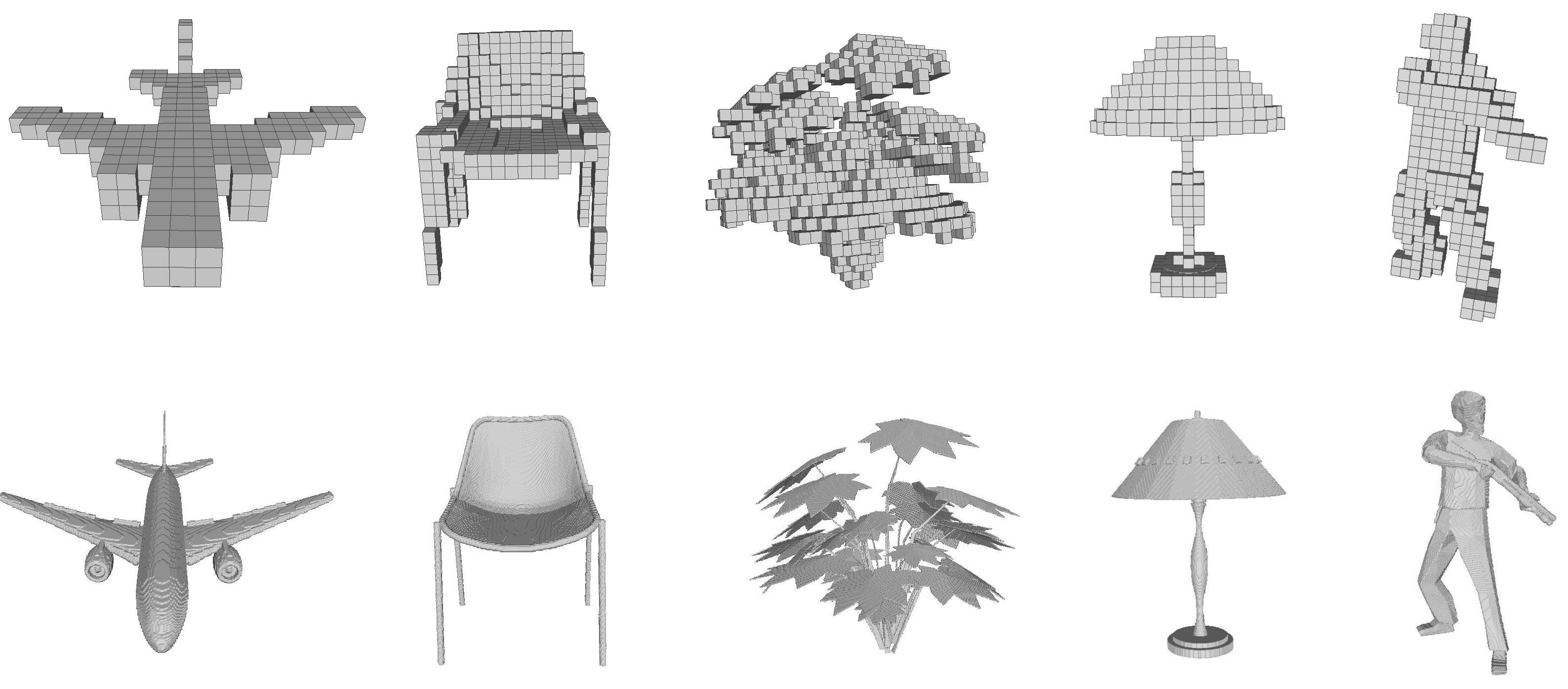}
\caption{Five example objects with the resolution $32^3$ in the top row and $512^3$ in the bottom row}\label{fig:differentResolution}
\centering
\end{figure}

Mesh voxelization is the process of converting a 3D triangular mesh into a 3D voxel grid. As we only use voxel models to compute SDF, resolution is no more a limitation to mesh voxelization, thus high quality shape representation with voxel model is available. In our work, we adopt solid mesh voxelization method in PyMesh\cite{pymesh} with the resolution of $512^3$. Fig.~\ref{fig:differentResolution} shows five example objects with the resolution of $32^3$ and $512^3$ separately. It can be seen that the voxel models with the resolution of $32^3$ lose fine details and still have a large number of voxels.

\subsection{SDF Computing}

After mesh voxelization, we calculate the SDF of each voxel whether it is inside or outside the object. Here we add a sphere with a radius of a half of voxelization resolution (256 in our setting) as an external boundary of the object.
As a result, the SDF value of a voxel outside the sphere will be empty. The SDF value of a voxel inside the object is negative while is positive if the corresponding voxel is outside the object but inside the external sphere.

\subsection{Infilling Sphere Construction}
Instead of using point cloud to represent a 3D model, our key idea is to fill the inside or outside of the 3D model with an appropriate number of infilling spheres. Here a sphere is defined by a voxel with the voxel coordinate as its center and the SDF value as its radius. Thus it is suitable to represent the spatial occupation of an object. However, the problem is how to construct a number of infilling spheres rather than using all spheres (voxels) to represent an object.

First, constructing infilling spheres should follow the principle of from big to small, from coarse to fine. By doing this, no matter how many spheres are used for object representation, larger spheres occupying the main part of a object as its basic trunk can be firstly constructed followed by filling the resting fine part with smaller spheres.

\begin{figure}[t]
\begin{center}
\includegraphics[width=0.7\textwidth]{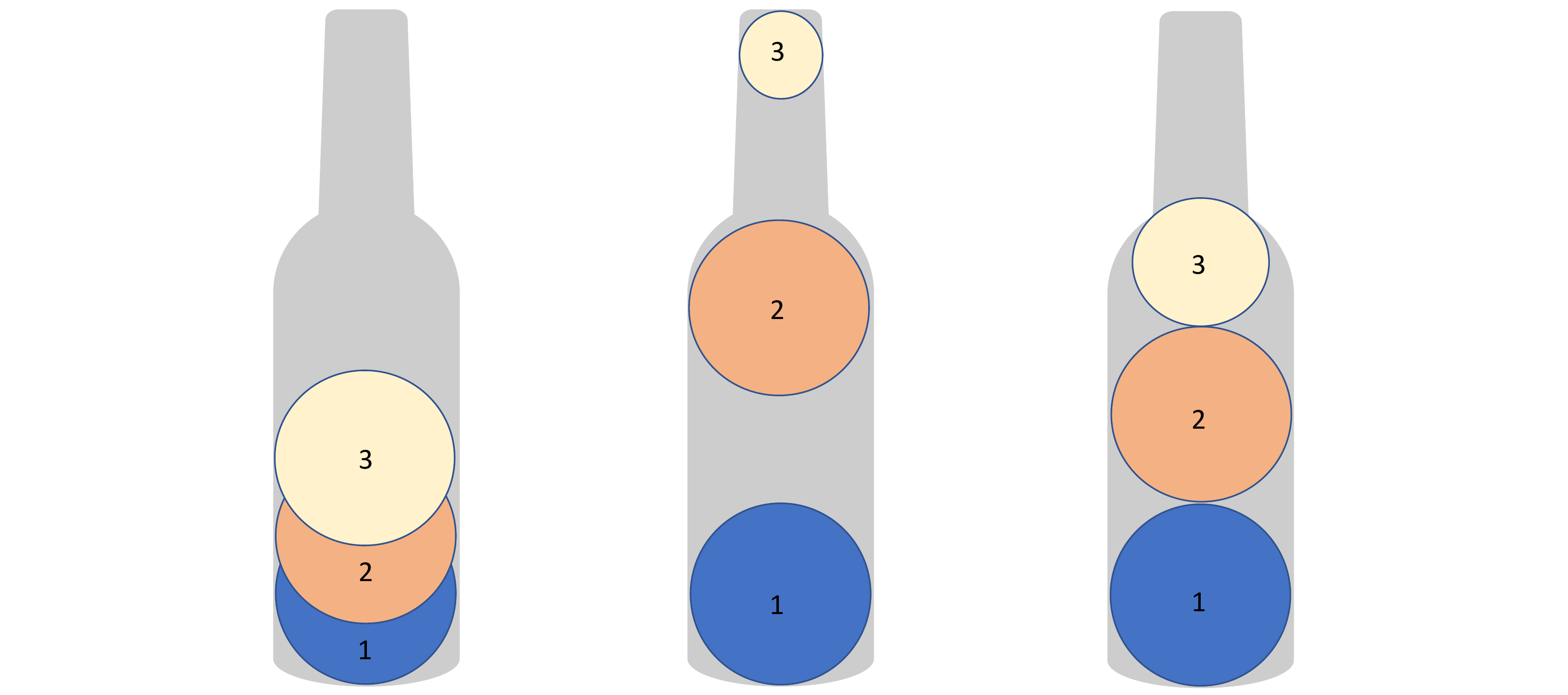}
\caption{The relationship of three largest spheres inside a bottle from left to right is: intersecting, separate, tangential.}
\label{fig:bottle_threeSpheres}
\end{center}
\end{figure}

\begin{figure}[!t]
\centering
\includegraphics[width=\textwidth]{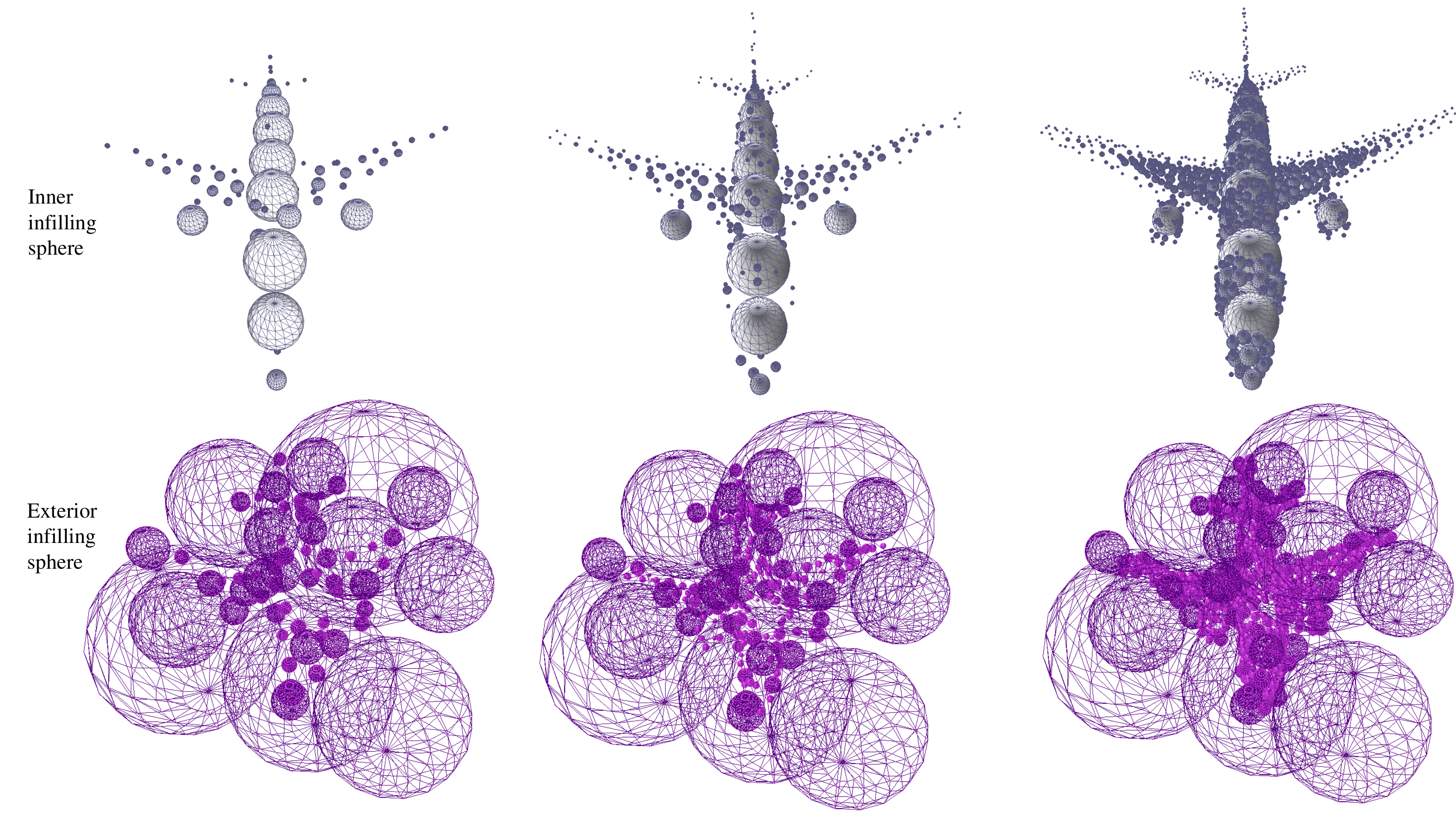}
\caption{Infilling spheres construction of an airplane model with different resolutions. The first and the second row draw inner and exterior infilling spheres respectively. The sphere resolutions $n$ from left to right are 64, 256, 1024 respectively.}
\label{fig:multi_resolution_InSpheres}
\centering
\end{figure}

Moreover, the adjacent relationship of infilling spheres should be considered. 
Take a bottle scenario as an example shown in Fig. \ref{fig:bottle_threeSpheres}. If given three infilling spheres to represent this bottle, the relationship of three largest infilling spheres could be classified as intersecting, tangential, and separate.
Obviously, the separate form should be adopted as the preferred occupation representation of the object because it is easier to occupy main space of the object at a given number $n$ of infilling spheres. 

Specifically, the ideal separate distance between two infilling spheres is affected by several factors, such as object volume, object shape, the number of infilling spheres, rendering the complicated constructing strategy. Here we finally adopt a simply hierarchical way which set the separate distance $d=10, 5, 0$ 
successively. As a result, any object can be represented hierarchically under a given infilling sphere resolution. Fig. \ref{fig:multi_resolution_InSpheres} shows the constructions of inner and exterior infilling spheres of an airplane model with different resolutions.

\begin{figure}[t]
\centering
\includegraphics[width=0.8\textwidth]{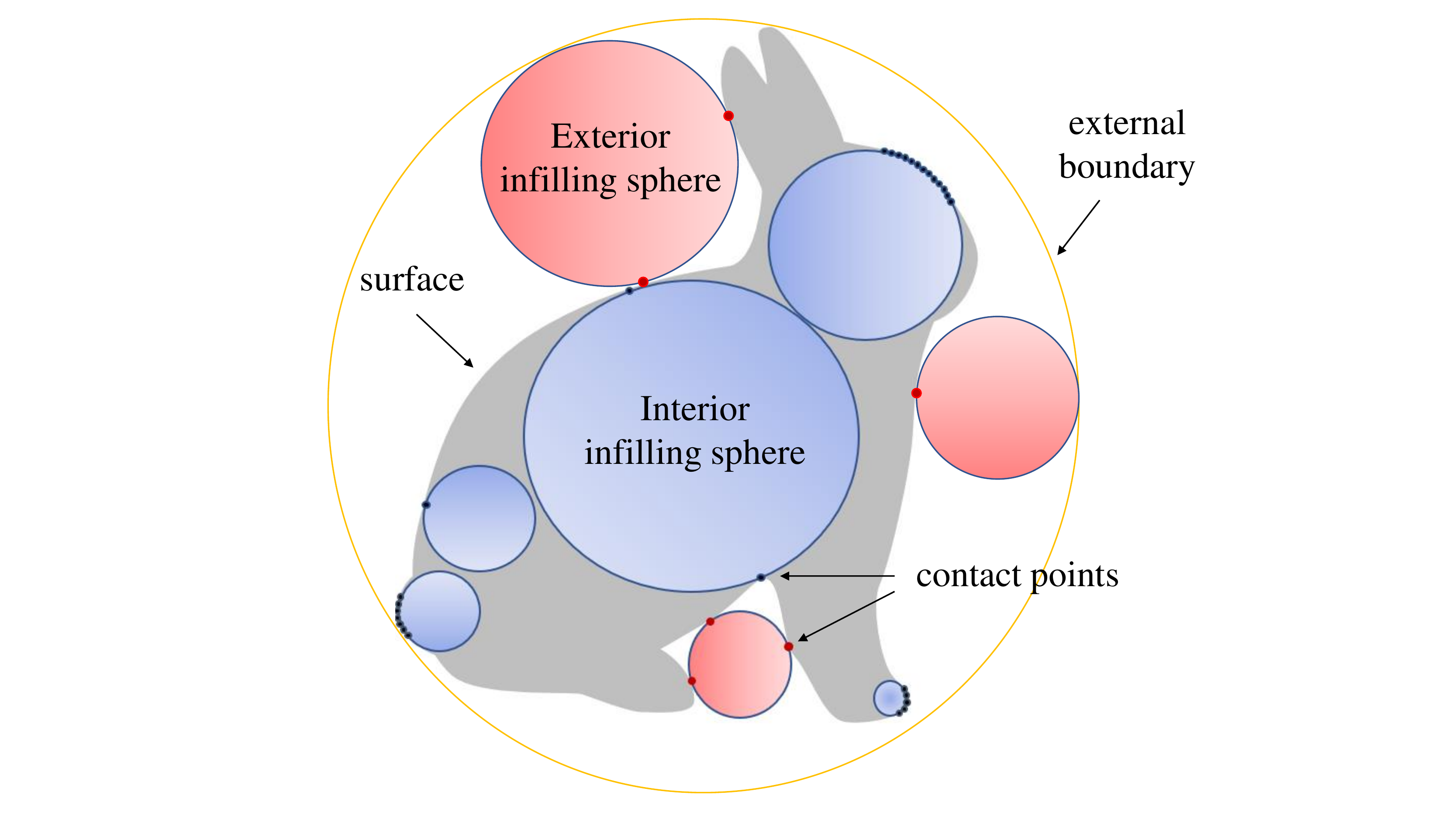}
\caption{Spheres with different numbers of contact points drawn in dark blue for interior case and red for exterior case}
\label{fig:different_contact_points}
\centering
\end{figure}

\begin{figure}[t]
\centering
\includegraphics[width=\textwidth]{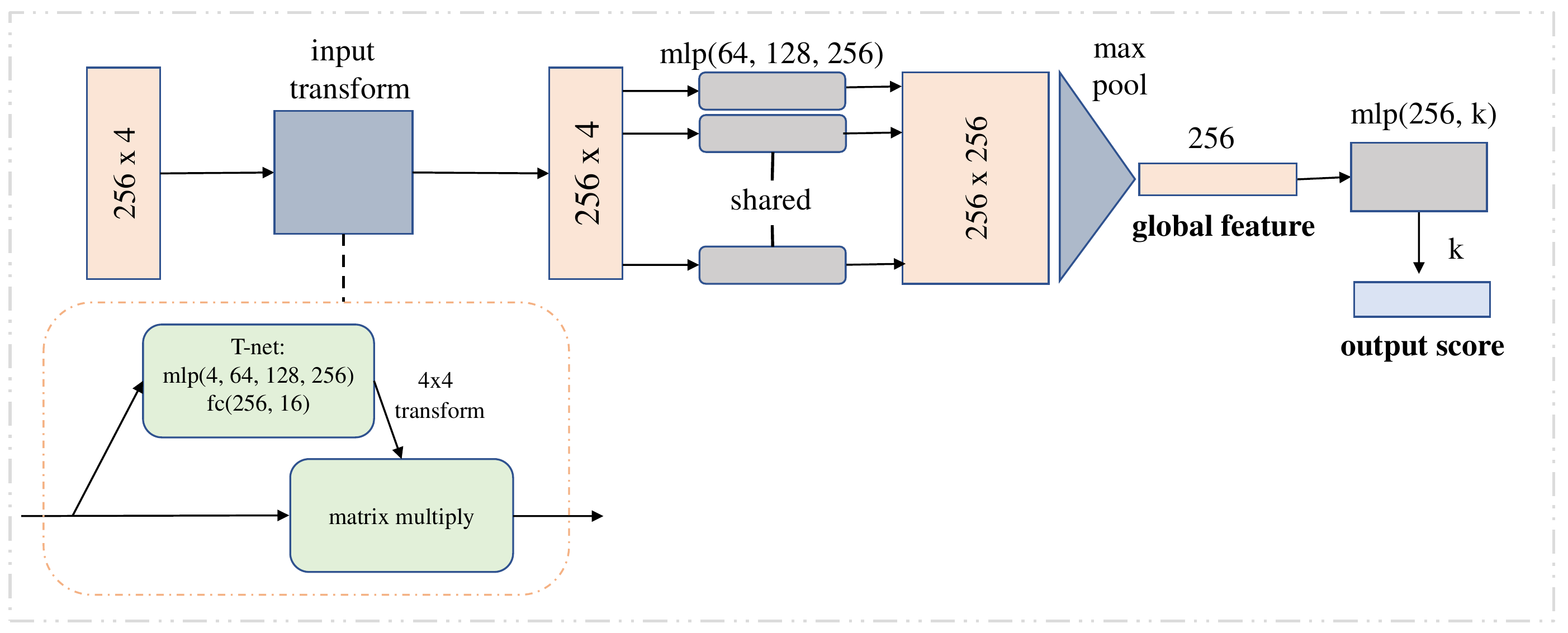}
\caption{Lightweight network architecture for 256 infilling spheres} 
\label{fig:lightweight_architecture}
\centering
\end{figure}

Another thing should be considered is that the selection of infilling spheres should also be related to the number of points contacting on the surface of the object. Fig. \ref{fig:different_contact_points} shows an example of a planar case. The sphere with more contacting points drawn in dark blue for interior case and red for exterior case usually includes richer shape information. In other words, such an infilling sphere is meaningful for representing the local shape of a 3D object. 

In summary, the three criteria of constructing infilling spheres are concluded as:
\begin{itemize}
    \item Infilling spheres should be constructed from large to small.
    \item Each infilling sphere should not intersect and try to avoid tangential with any other sphere.
    \item Each infilling sphere must be tangent to the object surface and has at least one contact point with the object.
 \end{itemize}

Note that we divide the infilling spheres construction into interior and exterior situations because both the geometric meaning and the sphere size for positive and negative SDF space are somewhat different. However, their processing flow is the same. The pseudo-code of interior/exterior infilling spheres construction is provided in Algo. 1 for a clear understanding.

\begin{algorithm}[!t]
\caption{Infilling spheres construction}
\label{alg:A}
\begin{algorithmic}
\STATE 0. Given infilling spheres number $n$; Let the number of constructed infilling spheres $m=0$.
\STATE 1. Mesh voxelization and SDF computation within an external sphere.
\STATE 2. Initialize status of each voxel ${V_i = empty}$.
\STATE 3. Obtain each sphere denoted by $X_i:=\{(x,s):SDF(x)=s\}$.
\STATE 4. Sort all $X_i$ with negative/positive values $s$ and the number of contacting points.
\STATE 5. Construct first infilling sphere $Y_1$. Let ${V_1 = true}$, $m=1$.
\STATE 6. Set $d=10$. \quad \quad \quad \% the distance threshold between different spheres
\STATE 7. for i = 2,3,4,...
\REPEAT
\STATE for j = 1-m
\REPEAT
\IF{ distance($x_i$, $x_j$)  $= d+s_j+s_i$  \quad \quad \quad \% $x_j$\: the center of one infilling sphere.}
\STATE construct a new infilling sphere $Y_{m+1}$, ${V_i = true}$, and $m=m+1$
\ELSE
\IF { distance($x_i$, $x_j$)  $<s_j+s_i$ \quad \quad \quad \% point $x_i$ is inside or near $Y_j$}
\STATE ${V_i = false}$
\ENDIF
\ENDIF
\UNTIL{maximum number of existing infilling spheres reached}
\UNTIL{maximum number of voxels or $n$ reached}
\STATE 8. Set $d=5$ and repeat Step 7 for voxel satisfying ${V_i = empty}$ until $n$ reached.
\STATE 9. Set $d=0$ and repeat Step 7 for voxel satisfying ${V_i = empty}$ until $n$ reached.
\end{algorithmic}
\end{algorithm}

\subsection{Neural Network Architecture}

When the infilling spheres are constructed, we can directly input these 4D primitives (coordinates of the sphere center plus the radius) into PointNet with a little adjustment to accomplish object classification.
However, since a few number of spheres are necessary for complete representation of a 3D object, we empirically propose a lightweight network architecture, leading to faster training and inference. The lightweight network architecture with the proposed infilling spheres as inputs are named \textbf{InSphereNet}. As demonstrated in experiments, InSphereNet can perform as well as PointNet even with only 12\% parameters and 17\% FLOPS. Fig. \ref{fig:lightweight_architecture} shows the designed lightweight network architecture for 256 infilling spheres. It is worth noting that there are only about 100000 parameters in this classification network.

\section{Experiment}
The experiments are divided into four parts. First, we show infilling spheres can be directly applied to classical classification network PointNet and achieve better performance with 1024 primitives. Second, to validate the robustness of the infilling spheres, we reduce the number of spheres to 512 and 256 separately. Although the number of input spheres decrease by $50\%$and $75\%$, the classification accuracy decrease slightly and become much better than PointNet with point clouds as input. Third, since the classification accuracy do not decrease significantly with less infilling sphere inputs, we reduce the dimension of global features and the number of fully connected layer, and the accuracy is still higher than $88\%$. Fourth, we visualize critical spheres which have the greatest impact on the classification task.

\subsection{Preliminary Classification Evaluation on ModelNet40}
PointNet learns global point cloud feature from 1024 uniformly sampled points of each 3D object in ModelNet40. To compare the representative ability of our spheres with point clouds, spheres with the same input data size are fed into the PointNet classification network. Table.  \ref{tab1} shows that three configurations of 1024 infilling spheres all achieve higher classification accuracy than 1024 points on PointNet. After that, We feed 1024 interior infilling spheres to PointNet++ without changing network architecture, the classification accuracy is also better than PointNet++. 

\begin{table}[t]
\caption{Overall classification accuracy on ModelNet40}\label{tab1} 
\centering
\begin{tabular}{llc}
\toprule
Method &  Input & Acc(\%)\\
\midrule
MVCNN 12x & images  & 89.5\\
3D Shapenets &  voxels  & 84.7\\
VoxNet & voxels & 85.9\\
PointNet & points  & 89.2\\
PointNet++ & points without normal & 90.7\\
PointNet++ & points with normal & 91.9\\
\midrule
Ours (1024 interior) & infilling spheres & 90.2\\
Ours (1024 exterior) & infilling spheres & 90.6\\
Ours (512 interior and 512 exterior) & infilling spheres & 90.3 \\
Ours (1024 interior on PointNet++) & infilling spheres & 92.1\\
\bottomrule
\end{tabular}
\centering
\end{table}

\subsection{Less Infilling Spheres Test}
\begin{figure}[t]
\centering
\includegraphics[width=0.7\textwidth]{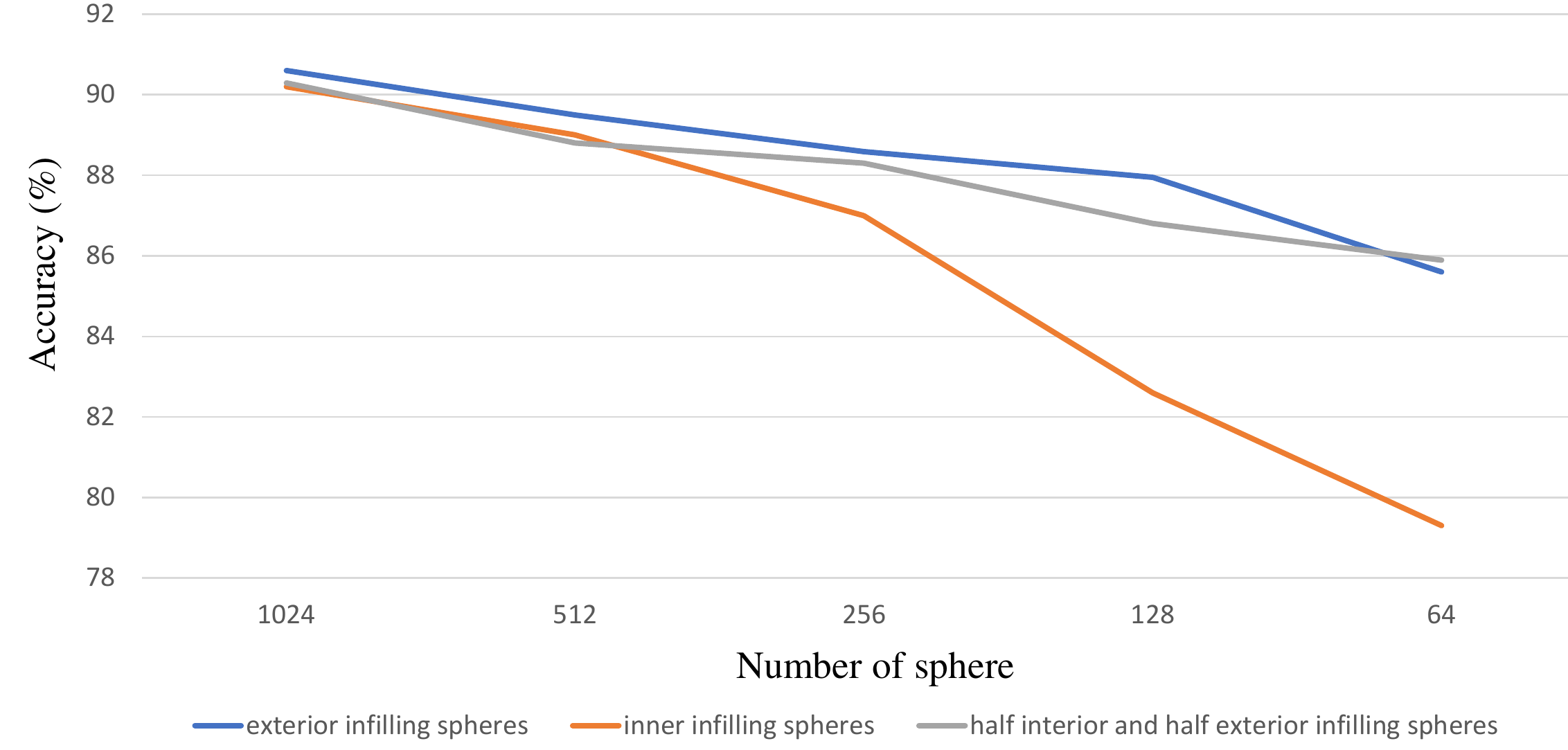}
\caption{Accuracy of InSphereNet vs. number of infilling spheres  }\label{fig:accuracy_VS_points_number}
\centering
\end{figure}

In this experiment, we show the infilling spheres for 3D classification task are more robust than point clouds in terms of the reduced network inputs. To point clouds, PointNet reports if there are $50\%$ points missing, the accuracy drops by $2.4\%$. However, when we reduce the number of infilling sphere inputs by 50\%, the accuracy only drop by $1.1\%$ for external infilling sphere and $1.2\%$ for inner infilling sphere separately. And the accuracy of 512 external infilling spheres is still higher than 1024 points. Fig. \ref{fig:accuracy_VS_points_number} depicts the overall classification accuracy with different numbers of infilling spheres for three configurations.

\begin{figure}[t]
\centering
\includegraphics[width=0.85\textwidth]{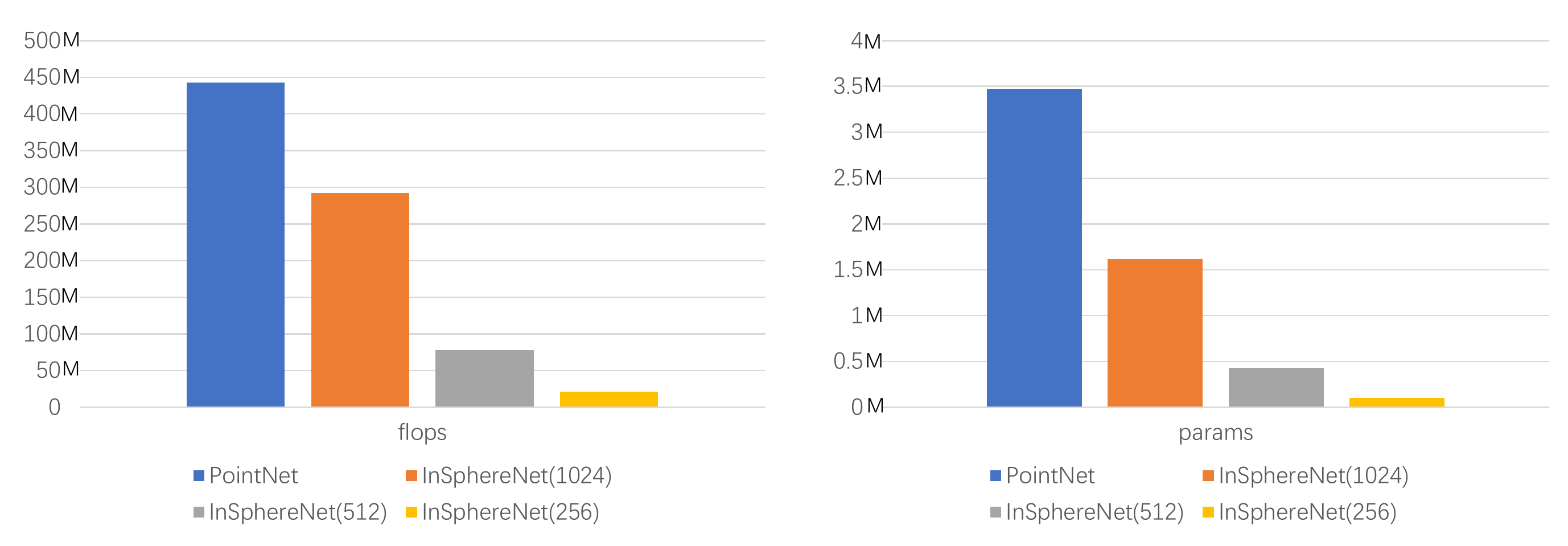}
\caption{Comparison of different network architectures in terms of flops and number of parameters} \label{fig:comparison_different_network}
\centering
\end{figure}

\subsection{Lightweight Neural Network Test}
From the above experiments, we found 512 external spheres still achieve higher classification accuracy than 1024 points. Therefore, we try modifying the network architecture settings to further test the effects of infilling spheres. As shown in Fig. \ref{fig:lightweight_architecture}, when the input size of external spheres is reduced from 1024 to 256, the dimension of SDF features extracted by the MLP layers is reduced accordingly from 1024 to 256. Since the feature dimension extracted by MLP decreases, we also reduce the fully connected layers, which largely reduced network parameters and computational costs. Fig. \ref{fig:comparison_different_network} shows the comparison of four different network architectures in terms of flops and number of parameters. Table. \ref{tab2} shows the classification accuracy for different network setting with different input data size. For 256 exterior spheres, the accuracy is even higher than $88\%$, which validates the representation power of infilling spheres.

\begin{table}[t]
\caption{Classification accuracy with different network setting}\label{tab2}
\centering
\begin{tabular}{lll}
\toprule
Input size & Network setting & Acc(\%) \\
\midrule
1024 & mlp(4,64,128,1024), fc(1024,512,256,k)&90.6\\
512 &mlp(4,64,128,512), fc(512,256,k) & 88.8\\
256 &mlp(4,64,128,256),fc(256,k) & 88.1\\
\bottomrule
\end{tabular}
\centering
\end{table}

\subsection{Critical Spheres Visualization}
\begin{figure}
\centering
\includegraphics[width=0.75\textwidth]{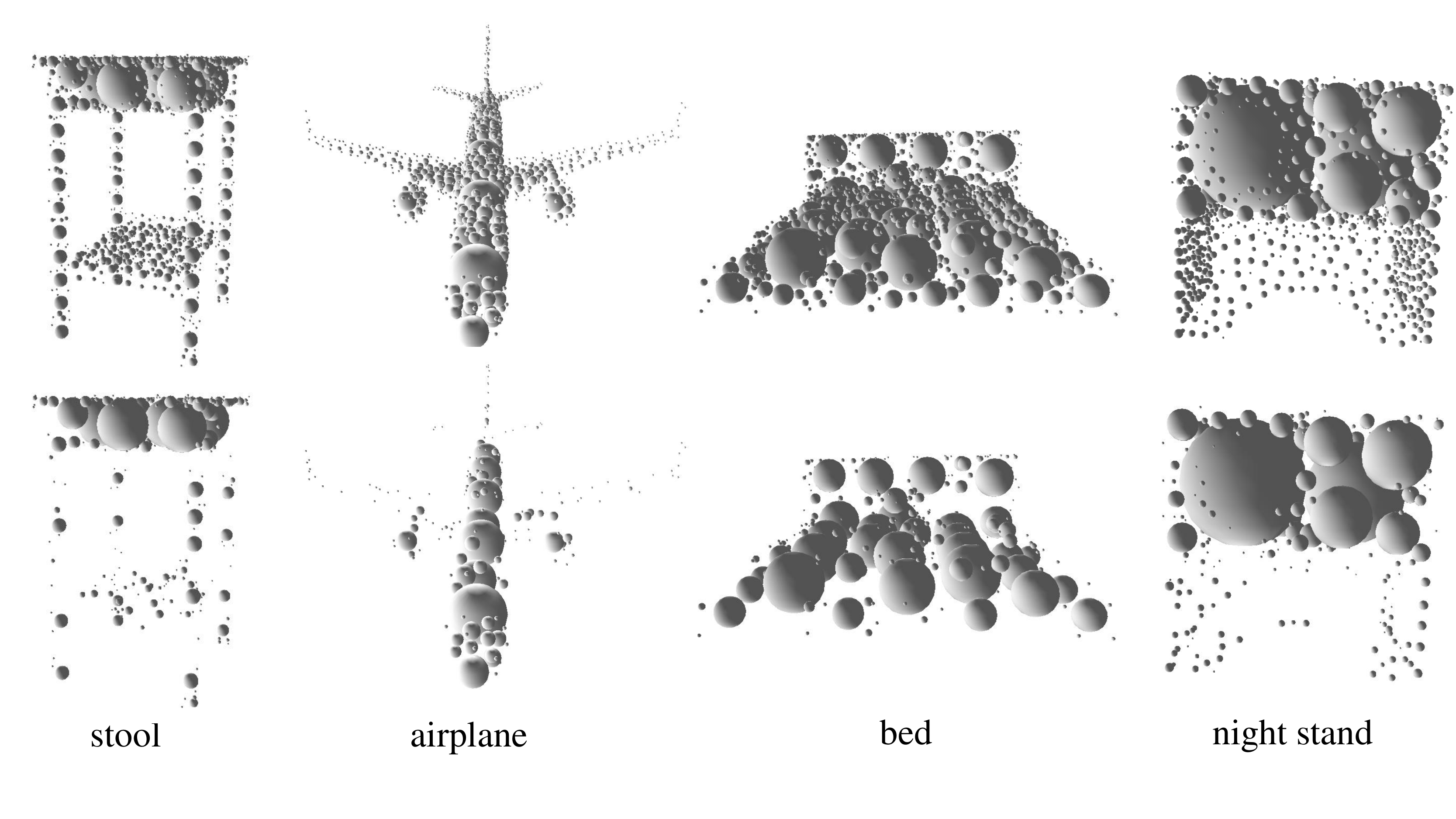}
\caption{Visualization of critical infilling spheres} \label{fig:visualization_critical_spheres}
\centering
\end{figure}

In this section, the critical infilling spheres with global feature are visualized. The original 1024 inputs are rendered in the first row of Fig. \ref{fig:visualization_critical_spheres} while the critical spheres are shown in its second row.
The number of critical spheres from left to right is only 251, 201, 267 and 234 respectively. It can be also seen that most of larger spheres are critical, which further proves the validity of our representation method for 3D objects. 

\section{Conclusion and Future Work}
In this paper, we present how to construct infilling spheres to accomplish 3D object classification. Compared to previous works directly utilizing point clouds on surface as inputs of DNN, the proposed method can represent 3D shape from coarse to fine as the number of infilling spheres increases. Experiment results show that InSphereNet has better performance than PointNet, especially with less number of inputting features. Even if the layers and parameters of DNN decreases sharply, the results are still satisfactory. All of this proves that infilling spheres are more representative and meaningful than point clouds.

One existing drawback of the proposed method is that the infilling spheres are still unstructured. In future work, we will try to extract local information of each infilling sphere by using graph convolution, just like PointNet++ and DGCNN do. 

\section*{Acknowledgements}

The authors would like to thank NSFC 61703092 for supporting this research.

\small
\bibliographystyle{splncs04}
\bibliography{samplepaper}
\end{document}